\documentclass[letter, 10 pt, conference]{ieeeconf}  

\IEEEoverridecommandlockouts 
\usepackage[english]{babel}
\usepackage[utf8]{inputenc}
\usepackage{times} 
\usepackage{cite}

\DeclareMathAlphabet{\pazocal}{OMS}{zplm}{m}{n}

\usepackage{hyperref}
\usepackage[nolist]{acronym}
\usepackage{balance} 

\usepackage{multirow}

\usepackage{siunitx}
\usepackage{graphicx} 
\usepackage{epsfig} 
\usepackage{pgfplots}
\usepackage{caption}
\usepackage[font=footnotesize]{subcaption}
\usepackage{pgfplots}

\usepackage{float}

\usepackage[export]{adjustbox}

\usepackage{xcolor}
\definecolor{myYellow}{rgb}{0.93,0.69,0.13}
\definecolor{myPurple}{rgb}{0.49,0.18,0.56}
\definecolor{myGreen}{rgb}{0.26 0.72 0.54}

\usepackage{mathtools}
\usepackage{nicefrac}
\usepackage{amsmath}
\usepackage{amssymb}
\usepackage{bm} 

\DeclareMathOperator*{\minimize}{minimize}

\usepackage{etoolbox}
\makeatletter%
\AfterPreamble{%
	\usepackage{hyperref}%
	\let\oldhypertarget\hypertarget%
	\renewcommand{\hypertarget}[2]{%
		\oldhypertarget{#1}{#2}%
		\protected@write\@mainaux{}{%
			\string\expandafter\string\gdef%
			\string\csname\string\detokenize{#1}\string\endcsname{#2}%
		}%
	}%
	\newcommand{\myhyperlink}[1]{%
		\hyperlink{#1}{\csname #1\endcsname}%
	}%
}
\makeatother%

\newcounter{Remark}
\newcounter{Definition}
\newcounter{Problem}
\usepackage{algorithm}
\usepackage[noend]{algpseudocode}

\makeatletter
\def\BState{\State\hskip-\ALG@thistlm}
\makeatother

\usepackage{tikz}
\pgfdeclarelayer{foreground}
\pgfsetlayers{background, main, foreground}
\usetikzlibrary{quotes, angles, backgrounds, arrows, automata, shapes, positioning, calc, through, spy, decorations.pathreplacing, decorations.markings, arrows.meta, automata, petri}

\tikzset{
    imglabel/.style={
      rectangle,
      inner sep=2pt,
      text=black,
      minimum height=1em,
      text centered,
      fill=white,
      fill opacity=1.0,
      text opacity=1,
      anchor=south west,
    },
  }
\tikzset{
	state/.style={
		rectangle,
		draw=black, very thick,
		minimum height=1.0em,
		text centered,
	},
}
\tikzset{
  on each segment/.style={
    decorate,
    decoration={
      show path construction,
      moveto code={},
      lineto code={
        \path [#1]
        (\tikzinputsegmentfirst) -- (\tikzinputsegmentlast);
      },
      curveto code={
        \path [#1] (\tikzinputsegmentfirst)
        .. controls
        (\tikzinputsegmentsupporta) and (\tikzinputsegmentsupportb)
        ..
        (\tikzinputsegmentlast);
      },
      closepath code={
        \path [#1]
        (\tikzinputsegmentfirst) -- (\tikzinputsegmentlast);
      },
    },
  },
  mid arrow/.style={postaction={decorate,decoration={
        markings,
        mark=at position .5 with {\arrow[#1]{stealth}}
      }}},
}

\graphicspath{{./figures/}}

\newcommand\copyrighttext{%
    \small \begin{center} \color{red} \textcopyright\,Accepted for presentation to the ``Beyond the Lab: Robust Planning and Control in Real World Scenarios" Workshop at ICRA'25, 19–23 May 2025, Atlanta, USA. Personal use of this material is permitted. Permission from authors must be obtained for all other uses, in any current or future media, including reprinting/republishing this material for advertising or promotional purposes, creating new collective works, for resale or redistribution to servers or lists, or reuse of any copyrighted component of this work in other works. \end{center}}
\newcommand\copyrightnotice{%
	\begin{tikzpicture}[remember picture,overlay]
	\node[anchor=south,yshift=25.6cm] at (current page.south) 
	{\color{red}\fbox{\parbox{\dimexpr\textwidth-\fboxsep-\fboxrule\relax}{\copyrighttext}}};
	\end{tikzpicture}%
}

\title{\copyrightnotice \LARGE \bf Robust Planning and Control of Omnidirectional MRAVs for Aerial Communications in Wireless Networks}

\author{Giuseppe Silano$^{1,2}$, Daniel Bonilla Licea$^{2,3}$, Hajar El Hammouti$^{3}$, Mounir Ghogho$^{3}$, and Martin Saska$^{2}$
    \thanks{$^1$Ricerca sul Sistema Energetico, Milan, Italy. 
    $^2$Czech Technical University in Prague, Prague, Czech Republic. 
    $^3$College of Computing, Mohammed VI Polytechnic University, Ben Guerir, Morocco.}
    \thanks{This work was partially funded by the EU under ROBOPROX (reg. no. CZ.02.01.01/00/22\_008/0004590), by the EU under the ARISE programme grant no. DCI-PANAF/2020/420-028, by the GAČR project no. 23-07517S, by the CTU grant no. SGS23/177/OHK3/3T/13, and by the research fund for the Italian Electrical System (Ricerca di Sistema) under the decree n. 388 of November 6th, 2024.}
} 

\begin{document}

\maketitle
\thispagestyle{empty} 
\pagestyle{empty} 


\begin{acronym}
    \acro{AoA}[AoA]{Angle of Arrival}
    \acro{AoD}[AoD]{Angle of Departure}
    \acro{BS}[BS]{Base Station}
    \acro{CoM}[CoM]{Center of Mass}
    \acro{DoF}[DoF]{Degrees of Freedom}
    \acro{FSO}[FSO]{Free-Space Optical}
    \acro{IoT}[IoT]{Internet of Thing}
    \acro{LoS}[LoS]{Line-of-Sight}
    \acro{MRAV}[MRAV]{Multi-Rotor Aerial Vehicle}
    \acro{NMPC}[NMPC]{Nonlinear Model Predictive Control}
    \acro{UAV}[UAV]{Unmanned Aerial Vehicle}
    \acro{f-MRAV}[f-MRAV]{fully actuated MRAV}
    \acro{u-MRAV}[u-MRAV]{under-actuated MRAV}
    \acro{o-MRAV}[o-MRAV]{omnidirectional MRAV}
    \acro{RF}[RF]{Radio Frequency}
    \acro{RSSI}{Received Signal strength Indicator}
    \acro{SNR}[SNR]{Signal-to-Noise Ratio}
    \acro{SINR}[SINR]{Signal-to-Interference-plus-Noise Ratio}
    \acro{wrt}[w.r.t.]{with respect to}
\end{acronym}



\begin{abstract}
    A new class of \acp{MRAV}, known as \acp{o-MRAV}, has gained attention for their ability to independently control 3D position and orientation. This capability enhances robust planning and control in aerial communication networks, enabling more adaptive trajectory planning and precise antenna alignment without additional mechanical components. These features are particularly valuable in uncertain environments, where disturbances such as wind and interference affect communication stability. This paper examines \acp{o-MRAV} in the context of robust aerial network planning, comparing them with the more common \acp{u-MRAV}. Key applications, including physical layer security, optical communications, and network densification, are highlighted, demonstrating the potential of \acp{o-MRAV} to improve reliability and efficiency in dynamic communication scenarios.
\end{abstract}



\section*{Full-version}
\label{sec:fullVersion}

The full-version of this paper is available at~\url{https://ieeexplore.ieee.org/document/10829762}. To reference, see~\cite{Licea2025IEEEComMag}.



\section{Introduction}
\label{sec:introduction}

The integration of \acfp{MRAV} into wireless networks has gained significant attention due to their agility, rapid deployment capabilities, and ability to establish line-of-sight communication. However, most \acp{UAV} used in communication systems are \acfp{u-MRAV}, which lack independent control over their 3D position and orientation. This limitation poses challenges in maintaining precise antenna alignment, especially for high-frequency technologies like millimeter-wave and terahertz communications, which require accurate beam alignment.

To overcome these challenges, \acfp{o-MRAV} have been introduced. Unlike traditional \acp{u-MRAV}, \acp{o-MRAV} can control both their position and orientation independently, enabling enhanced communication-aware trajectory planning and more reliable network performance. This capability is particularly advantageous for applications such as physical layer security, \ac{FSO} communications, and interference mitigation. Despite their advantages, \acp{o-MRAV} introduce additional complexity in terms of design, control strategies, and energy efficiency, which must be addressed for their practical deployment in real-world scenarios.

This paper explores the potential of \acp{o-MRAV} in enhancing aerial communication robustness in dynamic environments. It presents their unique capabilities compared to traditional \acp{UAV} and discusses the challenges and opportunities associated with their integration into modern communication networks.



\section{System Model and Control Capabilities}
\label{sec:systemModelControlCapabilities}

\acp{o-MRAV} introduce a novel control paradigm in \ac{MRAV}-enabled communication networks. Unlike \acp{u-MRAV}, which couple position and orientation, \acp{o-MRAV} possess full actuation, allowing independent 3D position and orientation control. This capability enables robust motion planning and precise antenna alignment, crucial for high-frequency wireless communication technologies such as mmWave and terahertz bands.

The actuation design of \acp{o-MRAV} relies on advanced tilting and bidirectional propeller mechanisms, enabling precise adjustments to compensate for disturbances. In contrast to \acp{u-MRAV}, which require movement to adjust antenna direction, \acp{o-MRAV} can maintain optimal alignment without altering their trajectory, ensuring consistent communication quality. These control advantages are particularly significant in scenarios requiring low-latency, high-reliability links, such as aerial base stations and disaster recovery networks.



\section{Aerial Communication Applications}
\label{sec:applications}

The unique control characteristics of \acp{o-MRAV} open new avenues for robust planning and optimization in wireless networks, enabling advancements in areas such as physical layer security, free-space optical communications, and network densification. These capabilities enhance communication reliability, interference mitigation, and adaptive network deployment, making \acp{o-MRAV} a valuable asset in dynamic and uncertain environments.

\textit{Physical Layer Security and Anti-Jamming.} 
Wireless network security is highly dependent on the radiation patterns of the antennas used by the legitimate nodes. Traditional physical layer security relies on multi-antenna systems to implement beamforming for interference suppression and improve secrecy of the communications. In contrast, \acp{o-MRAV} can physically reorient their onboard antennas to direct the nulls of the radiation patterns towards malicious nodes to neutralize them while maximizing the communications quality experienced by the legitimate users. This mechanical manipulation of the antenna radiation pattern orientation enhances secrecy rate and mitigates intentional interference, improving communication resilience under adversarial conditions.

\textit{\acl{FSO} Communications.} \ac{FSO} communication requires precise laser beam alignment between airborne nodes. Conventional \acp{MRAV} struggle with maintaining stable optical links due to orientation drift and aerodynamic disturbances. \acp{o-MRAV} improve link stability by allowing real-time orientation control, reducing jitter and ensuring continuous high-throughput optical connectivity. This capability is essential for low-latency, high-bandwidth applications in \ac{MRAV}-based relay networks.

\textit{Network Densification and Capacity Enhancement.} In urban environments, aerial base stations deployed via \acp{MRAV} must optimize coverage while minimizing interference. \acp{o-MRAV} offer fine-grained spatial control, allowing directional antenna steering for beam shaping and interference avoidance. This leads to better spectral efficiency in dense deployments and improves the adaptability of \ac{MRAV}-based networks in dynamic conditions.



\section{Case Study: Omnidirectional MRAV Relay}
\label{sec:caseStudy}

To illustrate the impact of \acp{o-MRAV} in robust communications, a case study is presented in which an \ac{o-MRAV} functions as an aerial relay between a mobile \ac{MRAV} and a ground-based \ac{BS}. This scenario highlights the advantages of independent position and orientation control in mitigating beam misalignment, improving network resilience, and ensuring consistent communication quality under dynamic conditions.

\begin{figure}[tb]
    \centering
    \includegraphics[scale=1]{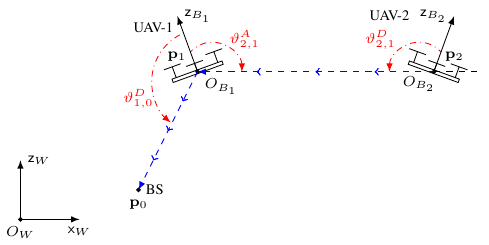}
	\vspace{-0.5em}
	\caption{Schematic of the communication relay scenario.}
	\label{fig:scenario}
\end{figure}

Consider a mobile \ac{MRAV} (\ac{UAV}-2) that must transmit data to a ground \ac{BS} while executing a mission that requires significant maneuvering. Due to its mobility constraints the high directionality of the antennas, maintaining a stable link to the \ac{BS} is challenging, especially when operating in environments where obstacles or long distances introduce signal degradation. To improve link quality, an omnidirectional \ac{MRAV} (\ac{UAV}-1) is introduced as a relay, dynamically adjusting its position and orientation to optimize signal transmission between \ac{UAV}-2 and the \ac{BS}. The scenario is schematically represented in Figure \ref{fig:scenario}.

To ensure real-time adaptability, a \ac{NMPC} strategy is designed, integrating both robotic dynamics and communication constraints into a predictive control framework. The \ac{NMPC} problem is formulated as an optimal control problem over a finite horizon N, solving for the control inputs $u$ that minimize communication misalignment and ensure stable transmission:
\begin{subequations}\label{eq:NMPC_formulation}
    \begin{align}
    &\minimize_{\bar{\mathbf{x}}, \, \bar{\mathbf{u}} } \;\;
    { \sum\limits_{k=0}^N \lVert \mathbf{y}_{\mathrm{d},k} - \mathbf{y}_k \rVert^2_{\mathbf{Q}} } \label{subeq:objectiveFunction} \\
    %
    &\quad \text{s.t.}~\; \bar{\mathbf{x}}_0 = \bar{\mathbf{x}}(\mathbf{t}_k), k = 0,  \label{subeq:stateEquation} \\
    &\;\;\; \quad \quad \bar{\mathbf{x}}_{k+1} = \mathbf{f}(\bar{\mathbf{x}}_k, \bar{\mathbf{u}}_k), k \in \{0, N-1\} , \label{subeq:sysDynamic} \\
    &\;\;\; \quad \quad \mathbf{y}_k = \mathbf{h}(\bar{\mathbf{x}}_k, \bar{\mathbf{u}}_k), k \in \{0, N\}, \label{subeq:outputMap} \\
    &\;\;\; \quad \quad\underline{\gamma} \leq \mathbf{u}_k \leq \bar{\gamma}, k \in \{0, N\}, \label{subeq:uBound} \\
    &\;\;\; \quad \quad \underline{\dot{\gamma}} \leq \bar{\mathbf{u}}_k \leq \bar{\dot{\gamma}}, k \in \{0, N-1\} \label{subeq:dotuBound}, \\
    &\;\;\; \quad \quad \mathbf{g}(\mathbf{u}_k, \mathbf{x}_k, \mathbf{y}_{\mathrm{d},k}, \mathcal{T}) > 0 \label{sueq:misAligConstr}, 
    \end{align}
\end{subequations}
where~\eqref{subeq:objectiveFunction} is the objective function,~\eqref{subeq:stateEquation} sets the initial state conditions,~\eqref{subeq:sysDynamic} and~\eqref{subeq:outputMap} express the discretized dynamic model for the~\ac{MRAV} and the output signals of the system, respectively, and actuator limits ($\underline{\gamma}, \bar{\gamma}, \underline{\dot{\gamma}}, \bar{\dot{\gamma}}$) are embedded in~\eqref{subeq:uBound} and~\eqref{subeq:dotuBound}. The constraints~\eqref{sueq:misAligConstr} ensure that~\ac{MRAV}-1 will be aligned to~\ac{MRAV}-2 and the~\ac{BS} while moving. The variable $\mathcal{T}$ refers to communication parameters that need to be taken into consideration while solving the problem. Finally, the vectors $\bar{\mathbf{u}}_k$, $\bar{\mathbf{x}}_k$, $\mathbf{y}_{\mathrm{d},k}$, and $\mathbf{y}_k$ denote the $k$-th element of vectors $\bar{\mathbf{u}}$, $\bar{\mathbf{x}}$, $\mathbf{y}_{\mathrm{d}}$, and $\mathbf{y}$, respectively. The feasibility and effectiveness of the control strategy have been demonstrated via closed-loop simulations in MATLAB, as discussed in \cite{BonillaICUAS2023}, but are not reported here due to space constraints.




\section{Challenges and Future Research Directions}
\label{sec:challengesResearchDirections}

Despite their advantages, several challenges must be addressed to fully integrate \acp{o-MRAV} into robust \ac{MRAV}-enabled networks. One major concern is \textit{energy efficiency}, as the added control capabilities increase power consumption, requiring optimized flight planning and power-aware actuation strategies. Another critical challenge lies in \textit{computational complexity}, since real-time trajectory and orientation optimization demand efficient algorithms capable of handling nonlinear dynamics and uncertainty propagation. Additionally, \textit{scalability in networked \acp{MRAV}} remains an issue, as coordinating multiple \acp{o-MRAV} in collaborative aerial communication introduces difficulties in synchronization, interference management, and distributed control. To overcome these challenges, future research should explore hybrid approaches that integrate mechanical and electronic beamforming, machine learning-driven adaptive control, and multi-agent coordination strategies, ensuring the effective deployment of \acp{o-MRAV} in aerial wireless networks.







\bibliographystyle{IEEEtran}
\bibliography{bib_short}

\end{document}